\documentclass{article}
\usepackage[preprint]{neurips_2019}

\usepackage[utf8]{inputenc} 
\usepackage[T1]{fontenc}    
\usepackage{hyperref}       
\usepackage{url}            
\usepackage{booktabs}       
\usepackage{amsfonts}       
\usepackage{nicefrac}       
\usepackage{microtype}      

\usepackage{helvet}  
\usepackage{courier}  
\usepackage{url}  
\usepackage{graphicx}  
\usepackage{algorithmic}
\usepackage{algorithm}
\usepackage{wrapfig}
\usepackage{lipsum}

\usepackage{times}
\usepackage{xspace}

\usepackage{graphicx} 
\usepackage{subfigure} 
\usepackage{caption}
\usepackage{paralist}
\usepackage{float}
\usepackage{tabularx}
\usepackage{multirow}
\usepackage{diagbox}
\usepackage{tikz}
\usepackage{booktabs}
\usepackage{array, makecell} 

\usepackage{natbib}
\usepackage{amsmath,amssymb}
\usepackage[inline]{enumitem}
\usepackage{acronym}

\DeclareMathOperator*{\argmax}{arg\,max}

\newcounter{todocnt}

\newcounter{latercnt}

\acrodef{RL}{Reinforcement Learning}
\acrodef{DRL}{Deep Reinforcement Learning}
\acrodef{IRL}{Inverse Reinforcement Learning}
\acrodef{SERP}{search engine result page}
\acrodef{IR}{Information Retrieval}
\acrodef{MDP}{Markov Decision Process}
\acrodef{MaxEnt-IRL}{Maximum Entropy Inverse Reinforcement Learning}
\acrodef{DM-IRL}{Distance Minimization Inverse Reinforcement Learning}
\acrodef{ISO}{Interactive System Optimizer}

\newcommand{\is}{interactive system\xspace}
\newcommand{\iss}{interactive systems\xspace}
\newcommand{\Iss}{Interactive systems\xspace}

\newcommand{\optimize}{Eq.~\ref{eq:optimal_environment}\xspace}

\newcommand{\statevalue}{Eq.~\ref{eq:state_value}\xspace}

\title{Learning Data-Driven Objectives to Optimize Interactive Systems}
\author{%
  Ziming Li\\
  University of Amsterdam\\
  The Netherlands \\
  \texttt{z.li@uva.nl} \\
  \And
  Julia Kiseleva\\
  Microsoft Research AI\\
  USA \\
  \texttt{jukisele@microsoft.com} \\
  \And
  Alekh Agarwal\\
  Microsoft Research AI\\
  USA \\
  \texttt{alekha@microsoft.com} \\
  \And
  Maarten de Rijke\\
  University of Amsterdam\\
  The Netherlands \\
  \texttt{M.deRijke@uva.nl} \\
}

\begin{document} 

\maketitle

\begin{abstract} 

Effective optimization is essential for \iss to provide a satisfactory user experience.
However, it is often challenging to find an objective to optimize for.
Generally, such objectives are manually crafted and rarely capture complex user needs in an accurate manner.
We propose an approach that infers the objective directly from observed user interactions.
These inferences can be made regardless of prior knowledge and across different types of user behavior.
We introduce \acf{ISO}, a novel algorithm that uses these inferred objectives for optimization.
Our main contribution is a new general principled approach to optimizing \iss using data-driven objectives.
We demonstrate the high effectiveness of \ac{ISO} over several simulations.

\end{abstract}

\section{Introduction}
\label{sec:intro}

\Iss play an important role in assisting users in a wide range of tasks~\citep{argelaguet2016role,kiseleva_sigir_2016,Li_emnlp_2016,Williams_www_2016}. 
They are characterized by repeated interactions with humans which follow the request-response schema, where the user takes an action first, and then the \is produces a reply. Such interactions can continue for several iterations until the user decides to stop when they are either satisfied or frustrated with their experience.
Importantly, an \is and its users always have a shared goal: for users to have the best experience.
But despite their shared goal, only the user can observe their own experience, leaving \iss unable to directly optimize their behavior. 

Currently, optimizing \iss relies on assumptions about user needs and frustrations~\citep{li2017towards}.
Commonly, an objective function is manually designed for a particular task to reflect the quality of an \is in terms of user satisfaction~\citep{kelly-methods-2009}.
The drawbacks of this approach are that a handcrafted objective function is heavily based on domain knowledge, that it is expensive to maintain, and that it does not generalize over different tasks.
Moreover, it is impossible to design such functions when there is a lack of domain knowledge.
%
Given an objective function, optimization can be done following the \ac{RL} paradigm; previous work does this by considering an \is as the agent and the underlying stochastic environment induced by the user~\citep{hofmann2013_wsdm, Li_emnlp_2016}.
However, user needs are inherently complex and depend on many different factors~\citep{Kosinski_2013, wei_wsdm_2017}. 
Consequently, manually crafted objective functions rarely correspond to the actual user experience. Therefore, even an \is that maximizes such an objective function is not expected to provide an optimal experience. In contrast with physically constrained environments~\citep{silver2016go, levine2016end2end, levine2016learning, finn_corl17}, the agent is supposed to adapt to a fixed environment that can not be changed. 
But users have difficulties in comprehending dramatic changes in an \is, so any changes should be done gradually to let users adapt to a newly optimized interactive system.

We propose a general perspective on how to overcome this discrepancy by simultaneously \begin{inparaenum}[(1)]\item inferring an objective function directly from data, namely user interactions with the system, and \item iteratively, and step by step, optimizing the system for this data-driven objective\end{inparaenum}.
In contrast to more traditional perspectives on optimizing interactive systems, our data-driven objectives are learned from user behavior, instead of being hand-crafted.
We hypothesize that by incorporating a data-driven objective, \iss can be optimized to an objective closer to the actual user satisfaction, unlike previous methods that optimize for assumed user preferences.

To this respect, we introduce a novel algorithm: \acf{ISO}.
It provides a principled approach by concurrently inferring data-driven objectives from user interactions and optimizing the \is accordingly. Thus, \ac{ISO} does not depend on any domain knowledge.

Below, we start by formalizing the interaction process between a user and an \is as a \ac{MDP} (Section~\ref{sec:interactions}).
Then we make the following contributions:
\begin{itemize}[nosep,leftmargin=*]
\item The first method that infers data-driven objectives solely from user interactions, that accurately reflect the users' needs without using any domain knowledge (Section~\ref{sec:objectives}).
\item A novel algorithm, \ac{ISO}, that optimizes an \is through data-driven objectives (Section~\ref{sec:opt-method}). 
Since users will change their behavior following updates of the \is, we devise experiments to simulate interactions in the real world. 
Our experiments with different types of simulated user behavior (Section~\ref{section:experimentalsetup}) show that \ac{ISO} increases the expected state value significantly.
\end{itemize}

\section{Related Work}
\label{section:relatedwork}

Relevant work for this paper comes in two broad strands: how to optimize interactive systems and what reward signal can be used for optimization.  

\textbf{Optimizing interactive systems.}
Interactive systems can be optimized by direct and indirect optimization. Direct optimization aims at maximizing the user satisfaction directly; in contrast, indirect optimization solves a related problem while hoping that its solution also maximizes user satisfaction \citep{dehghani2017neural}.
Direct optimization can be performed using supervised learning or \ac{RL}~\citep{mohri2012foundations}.
Many applications of \ac{RL} to optimizing interactive systems come from \ac{IR}, recommender systems, and dialogue systems. 
\citet{hofmann2011_ecir,hofmann-balancing-2013} apply \ac{RL} to optimize \ac{IR} systems; they use \ac{RL} for online learning to rank and use interleaving to infer user preferences~\citep{hofmann2013_wsdm}. 
\citet{shani2005mdp} describe an early MDP-based recommender system and report on its live deployment.
\citet{Li_emnlp_2016} apply \ac{RL} to optimize dialogue systems; in particular, they optimize handcrafted reward signals such as ease of answering, information flow, and semantic coherence. 

\textbf{Rewards for interactive systems.}
When applying \ac{RL} to the problem of optimizing interactive systems, we need to have rewards for at least some state-action pairs. 
Previous work typically handcrafts those, using, e.g., NDCG~\citep{odijk-dynamic-2015} or clicks \citep{kutlu2018correlation} before the optimization or the evaluation of the algorithm.
Instead of handcrafting rewards, we recover them from observed interactions between the user and the interactive system using \ac{IRL}.
\citet{ziebart_iui_2012} use \ac{IRL} for predicting the desired target of a partial pointing motion in graphical user interfaces.
\citet{monfort_aaai_2015} use \ac{IRL} to predict human motion when interacting with the environment. 
\ac{IRL} has also been applied to dialogues to extract the reward function and model the user \citep{pietquin2013inverse}. 
\ac{IRL} is used to model user behavior in order to make predictions about it. But
we use \ac{IRL} as a way to recover the rewards from user behavior instead of handcrafting them and optimize an interactive system using these recovered rewards.
The closest work to ours is by \citep{lowe2017towards}, who learn a function to evaluate dialogue responses. However, the authors stop at evaluation and do not actually optimize the interactive system.

The key difference between our work and previous studies is that we first use recovered rewards from observed user interactions to reflect user needs and define \is objectives. Subsequently the \is can be optimized according to the defined data-driven objectives so as to improve the user experience.

\section{Background}
\label{sec:background}

\acf{RL} and \acf{IRL} are the fundamental techniques used in the framework we propose in this paper.

In \ac{RL} an agent learns to alter its behavior through trial-and-error interactions with its environment~\citep{sutton1998rl}. 
The goal of the agent is to learn a policy that maximizes the expected return. 
\ac{RL} algorithms have successfully been applied to areas ranging from traditional games to robotics~\citep{mnih2015human,silver2016go,levine2016end2end,levine2016learning,duan2016rl,wang2016learning,zhu2017target}.

The task of \ac{IRL} is to extract a reward function given observed, optimal (or suboptimal) behavior of an agent over time~\citep{ng_icml_2000}. 
The main motivation behind \ac{IRL} is that designing an appropriate reward function for most \ac{RL} problems is non-trivial; this includes animal and human behavior~\citep{abbeel_icml_2004}, where the reward function is generally assumed to be fixed and can only be ascertained through empirical investigation. 
Thus inferring the reward function from historical behavior generated by an agent's policy can be an effective approach.
Another motivation comes from imitation learning, where the aim is to teach an agent to behave like an \emph{expert} agent. 
Instead of directly learning the agent's policy, other work first recovers the expert's reward function and then uses it to generate a policy that maximizes the expected accrued reward~\citep{ng_icml_2000}.
Since the inception of \ac{IRL}~\citep{Russell_1998aa}, several  IRL algorithms have been proposed, including maximum margin approaches~\citep{abbeel_icml_2004,ratliff2009learning},
and probabilistic approaches such as ~\citep{ziebart_aaai_2008} and \citep{boularias2011relative}. In the last few years,  a number of adversarial IRL methods \citep{finn2016connection,finn2016guided,ho2016generative,fu2017learning} have been proposed because of its ability to adapt training samples to improve learning efficiency.

\section{Modeling User-System Interactions}
\label{sec:interactions}

We assume that the agent is a user who interacts with the environment, an \is, 
with the goal of maximizing their expected rewards. As a running example we can consider a user who is interacting with a search engine.
This process is modeled using a finite \ac{MDP} $(S, A, T, r, \gamma)$, in the following way:
\begin{itemize}[nosep,leftmargin=*]
\item $S$ is a set of states that represent responses from the \is to the user. $S$ is finite as there are limited predefined number of responses that \is can return.
\item $A$ is a finite set of actions that the user can perform on the system to move between states. In case of search engine, a user can run a query, click on the returned results, reformulate a query etc.
\item $T$ is a transition distribution and $T(s, a, s')$ is the probability of transitioning from state $s$ to state $s'$ under action $a$ at time $t$: 
\begin{equation}
    T (s' \mid s, a) =  \mathbb P(S_{t+1}=s' \mid S_t = s, A_t=a).
\end{equation}
For search engines, being at the start page (which is $s$) a user is making an action $a$, e.g.\ running a query, and the engine redirects him to a result page (which is $s'$).

\item $r(s,a,s')$  is the expected immediate reward after transitioning from $s$ to $s'$ by taking action $a$. 
We compute the expected rewards for (state, action, next state) triples as:
\begin{equation}
\mbox{}\!\!
r(s,a,s') =  \mathbb{E}[R_t \mid S_t = s, A_t=a,  S_{t+1}=s'],    
\end{equation}
where $R_t$ is reward at time $t$. In case of search engine, a user is getting a reward for finding a desired information. However, the rewards are not observed.
For simplicity in exposition, we write rewards as $r(s)$ rather than $r(s, a, s')$ in our setting; the extension is trivial~\citep{ng_icml_2000}.
\item $\gamma \in [0,1]$ is a discount factor.
\end{itemize}

We write $\mathcal{P}$ to denote the set of interactive systems, i.e., triples of the form $(S,A,T)$. 
System designers have control over the sets $S$, $A$, and the transition distribution, $T$, and $T$ can be changed to optimize an \is. 

The \emph{user behavior strategy} is represented by a policy, which is a mapping, $\pi \in \Pi$, from states, $s \in S$, and actions, $a \in A$, to $\pi(a|s)$, which is the probability of performing action $A_t=a$ by the user when in state $S_t=s$. The observed history of interactions between the user and the \is, $H$, is represented as a set of trajectories, $\{\zeta_i\}_{i=1}^n$, drawn from a distribution $Z$, which is brought about by $T$, $\pi$, and $D_0$, where $D_0$ is the initial distribution of states. To simplify the problem, we assume that user behavior is homogeneous, i.e. one user generate $H$. 
A \emph{trajectory} is a sequence of state-action pairs:
\begin{equation}
\label{eq:traj}
\zeta_i = S_0,A_0, S_1,A_1, \dots, S_t, A_t, \dots. 
\end{equation}
In this paper we act under assumption that the user is an optimal agent in terms of maximizing his reward under the system dynamics, and the system wants to improve the user experience over time by creating progressively easier MDPs to solve for the user. However, a \is cannot transition from all initial to $s_0$ in one step, due to design constraints. For example, if a user is searching for holidays destination, the system cannot redirect him to the final stage of booking a hotel because he needs to go through the necessary step, e.g.\ paying for it.

\section{Defining Data-driven Objectives}
\label{sec:objectives}
\noindent
\textbf{Defining Interactive System Objectives.}
We define the \emph{quality} of an \is as the expected state value under an optimal user policy. The value of a state $S_0$ under a policy $\pi$ is given as:

\begin{equation}
  \upsilon_{\pi} (S_0) =  \mathbb{E}_{\pi} \left[\sum^{\infty}_{t=0} \gamma^t R_{t+1}\right],
\end{equation} 
where the expectation $\mathbb{E}_{\pi}[\cdot]$ is taken with respect to sequences of states $S_0, S_1, \dots, S_t, \dots$ drawn from the policy $\pi$ and transition distribution $T$. 
The quality of the \is under user policy $\pi$ is: 
\begin{equation}
\label{eq:state_value}
\mathbb{E}_{S_0 \sim D_0}[\upsilon_\pi(S_0)],
\end{equation}
where $D_0$ is the initial distribution of states.
In the proposed setting, the user goal is to find the best policy such that  $\mathbb{E}_{S_0 \sim D_0}[\upsilon_\pi(S_0)]$ is maximized.
$\upsilon_{*}(S_0)$ defines the maximum possible value of $\upsilon_{\pi }(S_0)$ as follows:
\begin{equation}
\upsilon_{*}(S_0)=\max_{\pi \in \Pi} \limits \upsilon_\pi(S_0), 
\end{equation}
where $\Pi$ is the set of possible user policies.
We formulate the problem of finding the optimal \is's transition distribution, denoted $T^*$, in the following terms:
\begin{equation}
\label{eq:optimal_environment}
T^* = \argmax _{T \in T}\mathbb{E}_{S_0 \sim D_0}[\upsilon_{*}(S_0)].
\end{equation}
Therefore, \optimize represents the objective that we use to optimize an \is in order to improve the user experience. 
To estimate these \is objectives, we first need to recover $R_t$, which we will discuss next. 

\noindent
\textbf{Recovering user rewards.}
We assume that continued user interactions with the system indicate a certain level of user satisfaction, which can be reflected by experienced rewards.
In contrast with $\zeta_i \in H$ presented in Eq.~\ref{eq:traj}, the complete history of interactions, $\hat{H}$, consists of trajectories $\hat{\zeta_i} \sim \hat{Z}$, which include the user reward $R_t$:
\begin{equation}
\label{eq:scored_trajectory}
\hat{\zeta_i} = S_0,A_0,R_1, S_1,A_1,R_2 \dots, R_t, S_t, A_t, \dots.
\end{equation}

The problem is that the true reward function is hidden and we need to recover it from the collected incomplete user trajectories, $H$, shown in Eq.~\ref{eq:traj}. 
To address this challenge we apply \ac{IRL} methods (Section~\ref{sec:background}), which are proposed to recover the rewards of different states, $r(s)$, for $\zeta_i \in H$.

Our assumptions about the user reward function is:
given state feature functions $\phi : {S_t} \to \mathbb{R}^k$ that describes $S_t$ as a $k$-dimensional feature vector, the true reward function $r(s)$ is a linear combination of the state features $\phi(s)$, which can be given as $r(s) = \theta^T \phi(s)$.

To uncover the reward weights $\theta$, we use \ac{MaxEnt-IRL}~\citep{ziebart_aaai_2008}, where the core idea is that trajectories with equivalent rewards have equal probability to be selected and trajectories with higher rewards are exponentially more preferred, which can be formulated as: 
%
$\mathbb{P}(\zeta_i\mid \theta) = \frac{1}{\Omega(\theta)} e^{\theta^T\phi({\zeta_i})} = \frac{1}{\Omega(\theta)} e^{\sum_{t=0}^{|\zeta_i|-1}\theta^T\phi(S_t)}$,
%
where $\Omega(\theta)$ is the partition function. 
\ac{MaxEnt-IRL} maximizes the likelihood of the observed data under the maximum entropy (exponential family) distribution. 
Once we have recovered the reward function $r(s)$ we can proceed to the optimization objectives presented in~\optimize.

\noindent
\textbf{Oracle rewards.}
For comparison, we employ \ac{DM-IRL}~\citep{el2013_dm_irl, Burchfiel_aaai_2016} to have a perfectly recovered reward weight $\theta$.
\ac{DM-IRL} directly attempts to regress the user's actual reward function that explains the given labels.
\ac{DM-IRL} uses discounted accrued features to represent the trajectory:
%
$\psi(\zeta_i) = \sum_{t=0}^{|\zeta_i|-1} \gamma^t \phi(S_t)$,
%
where $\gamma$ is the discount factor. The score of a trajectory $\zeta_i$ is 
%
$\text{score}_{\zeta_i} = \theta^T \psi(\zeta_i)$.
%
Since the exact score for each trajectory is supplied, the recovered rewards with \ac{DM-IRL} are exactly the ground truth of reward functions, which can be regarded as oracle rewards. 

\section{Optimizing Interactive Systems}
\label{sec:opt-method}
\begin{algorithm}
   \caption{Interactive System Optimizer (ISO)}
   \label{alg:opt-env}
   \begin{algorithmic}[1]
   \STATE {\bf Input:} Original system $(S, A, T)$, $r$, $\gamma$, $D_0$.
   \STATE {\bfseries Output:} Optimized system $(S, A, T^*)$ 
   \STATE  Construct original $\text{MDP}$$(S, A, T, r, \gamma)$
   \STATE  $\pi_*(a|s) = RL (S, A, T, r, \gamma)$
   \STATE Construct syst. MDP$^+(S^+, A^+, T^+, r^+, \gamma^+)$:
               \begin{itemize}[nosep]
               \item $S_t^+ = ( S_t, A_t)$
               \item $A_t^+ = S_{t+1}$
               \item $T^+ (S_{t+1}^+ | S_t^+, A_t^+)  = 
                          \pi_*(A_{t+1} | S_{t+1}) $
               \item $r(S_t^+)^+$ = $r(S_t)$
               \item $\gamma^+$ = $\gamma$
             \end{itemize}
   \STATE $D_0^+ \sim ( S_0 \sim D_0 , A_0 \sim \pi_*(a| S_0))$
   \STATE $\pi^+(A_t^+|S_t^+) = T(S_{t+1}|S_t, A_t)$         
   \STATE $\pi^{+}_*(a^+|s^+)  = RL (S^+, A^+, T^+, r^+, \gamma^+)$          
   \STATE $T^*(S_{t+1} | S_t, A_t)  = \pi^{+}_* (A_t^+ | S_t^+) $
\end{algorithmic}
\end{algorithm}

We start by explaining how to maximize the quality of an \is for a user behaving according to a fixed stationary policy $\pi$:
\begin{equation}
\label{eq:optimal_environment_fixed_policy}
T_{\pi}^* = \argmax _{T \in T}\mathbb{E}_{S_0 \sim D_0}[\upsilon_{\pi}(S_0)].
\end{equation}
This problem is equivalent to finding the optimal policy in a new $\text{MDP}^+(S^+,A^+, T^+,  r^+,\gamma^+)$, where the agent is an \is and the stochastic environment is a user. 
In $\text{MDP}^+$, the state $S_t^+ $ is represented by a combination of the state $S_t$ the user is in and the action $A_t$ the user takes at time step $t$ from the original MDP; the action $A^+_t$ is the original state $S_{t+1}$.
The \is observes the current state $S^+_t$ and picks an action $A^+_t$ under the \is policy $\pi^+(A^+_t|  S^+_t)$. Then the user returns the next state $S_{t+1}^{+}$ according to the transition distribution $T^+ (S_{t+1}^+ | S_t^+, A_t^+)$ which is inferred from the policy model $\pi (A_{t+1}  |  S_{t+1})$. 
Therefore, finding the optimal $T^{*}_{\pi}$ from Eq.~\ref{eq:optimal_environment_fixed_policy} is equivalent to finding the optimal $\pi_{*}^+$ for $\text{MDP}^+$ as follows:
\begin{equation}
\label{eq:optimal_policy_plus}
  \pi^{+}_{*} = \argmax _{\pi^+ \in \Pi^+} \mathbb{E}_{S_0^+ \sim D_0^+} [\upsilon_{\pi^+} (S_0^+)],
  \end{equation}

which can be done using an appropriate \ac{RL} method such as Q-learning or Policy Gradient. $D_0^+$ is the initial distribution of states in $\text{MDP}^+$.
After we have demonstrated how to optimize the \is for a given stationary policy, we return to the original problem of optimizing the \is for an optimal policy $\pi_*$.

We propose a procedure \ac{ISO} that is presented in Algorithm~\ref{alg:opt-env} and has the following main steps:

\emph{Line 1:} We assume that we have an estimate of the reward function $r(s)$ using one of the \ac{IRL} methods described in Section \ref{sec:interactions}, so we have as input: the original system $(S, A, T)$, the reward function $r$, the discount factor $\gamma$, and the initial distribution of states $D_0$.\\
\emph{Line 2:}  \ac{ISO} outputs the optimized \is $(S, A, T^*)$. \\
\noindent
\emph{Line 3:} \ac{ISO} formulates the original system as $\text{MDP}$$(S, A, T, r, \gamma)$.\\
\emph{Line 4:} \ac{ISO} uses an appropriate \ac{RL} algorithm to find the current policy $\pi_*(a|s)$ given the reward function $r$.  
\emph{Line 5:} \ac{ISO} transforms the original MDP$(S, A, T, r, \gamma)$ into the new $\text{MDP}^+(S^+,A^+, T^+ , r^+, \gamma^+)$. In our setting, $S^+_t$ has the same reward value as $S_t$. 

The discount factor $\gamma^+$ remains the same.\\
\emph{Line 6:} 
\ac{ISO} transforms $D_0$ to $D_0^+$ to match the distribution of first state-action pairs in the original MDP.\\
\emph{Line 7:} The equivalence $\pi^+(A_t^+ |  S_t^+) = T(S_{t+1} |  A_t,S_t)$ means that finding the optimal $\pi^+_{*}$ according to Eq.~\ref{eq:optimal_policy_plus} is equivalent to finding the optimal $T^*_{\pi}$ according to Eq.~\ref{eq:optimal_environment_fixed_policy}. Therefore, the transition distribution can be regarded as a policy network or a policy table from $\text{MDP}'$s perspective depending on the policy learning method.\\
\emph{Line 8:}
We can use an appropriate \ac{RL} algorithm to find $\pi_*^+(A_t^+ |  S_t^+)$.\\
\emph{Line 9:}
\ac{ISO} extracts $T^*(S_{t+1} |  S_t, A_t)$ from the optimal system policy $\pi_*^+(A_t^+ |  S_t^+)$. The extraction process is trivial: $T^*(S_{t+1} |  S_t, A_t) =  \pi^+(A_t^+ |  S_t^+)$. Therefore, \ac{ISO} terminates by returning the \emph{optimized} \is.

Once \ac{ISO} has delivered the optimized system $(S, A, T^{*})$, we expose it to users so they can interact with it. 
We assume that users adjust their policy to $T^{*}$. 
After enough iterations the user policy will converge to the optimal one.
Iterations between optimizing the \is for the current policy and updating the user policy for the current \is continue until both converge.

\section{Experimental Setup}
\label{section:experimentalsetup}

\noindent
\textbf{Designing an interactive system.}
For experimental evaluation we simulate an arbitrary \is where we need a finite set of states $S$, a finite set of actions $A$ and a transition distribution $T$. We use simulations as we require to have a changeable environment which is provided to users for interactions. \ac{ISO} is designed to improve an \is gradually (not in go).
Features of a state $\phi(s)$ are fixed. 
For our experimental setup, we simulate the \is where $|S| = 64$ and $|A| = 4$. 
We work with a complex environment where a user can transition between any two states if these two states are connected. 
The connections between two states are predefined and fixed, but the transition distribution is changeable. In another word, for the same system in different runs, the connectivity graph of this systems is fixed and will not be changed once it is sampled at the very beginning. 
We use a hyper-parameter, the connection factor $cf$, to define the number of possible next states after the user has taken one specific action at the current state.
For an initial \is, $D_0$ is randomly sampled as well as $T$. At each iteration \ac{ISO} delivers $T^*$, which substitutes the initial $T$. 

\noindent
\textbf{Modeling user behavior.}
To model user behavior we require a true reward function $r(s)$, and an optimal user policy $\pi_{*}$. 
We utilize a linear reward function $r(s)$ by randomly assigning $25\%$ of the states reward 1, while all others receive 0. As we use one-hot features for each state, $r(s)$ is guaranteed to be linear. 

We use a soft value iteration method~\citep{ziebart2010modeling} to obtain the optimal user policy $\pi^*$.

The quality of the recovered reward functions user behavior is influenced by how trajectories are created, which in turn can affect the performance of \ac{ISO} as it relies on $r(s)$ to optimize an \is. We experiment with the following types of user trajectories:
\begin{description}[nosep,leftmargin=*]
\item[\emph{Optimal:}] Users know how to behave optimally in an interactive system to satisfy their needs. To simulate it, we use $\pi_{*}$ trained with the real reward function.
\item[\emph{SubOptimal:}]

Not all users know the system well, which means that the demonstrated behavior is a mixture of optimal and random. We propose two different methods to get suboptimal behavior. The degree of optimality of user behavior is controlled by (1) the proportion of random behavior\footnote{To model suboptimal user behavior we use two user policies: (1)~an optimal user policy $\pi_{*}$; and (2)~an adversarial policy ($1-\pi_{*}$). We include an adversarial policy instead of a random one because it is the hardest case as users behave opposite of what we expect.} (2) or the user action noise, which are collectively called the noise factor (NF) $\in [0.0, 1.0]$.
\begin{description}
\item[Mix of Behaviors (MB):] The final dataset $H$ is a mix of trajectories generated by the optimal policy and the adversarial policy\footnote{E.g., $\mathit{NF} = 0.2$ means that 20\% of the trajectories are generated with the adversarial policy.}.
\item[Noise in Behavior (NB):] In this case, the trajectories in $H$ are generated from the optimal policy but we add noise to the user actions to get suboptimal behavior \footnote{E.g., $\mathit{NF} = 0.2$ means the probability is 20\% that the user will not choose the action with the highest probability in the optimal policy.}. 
\end{description}
\end{description}

The generated history of user interactions $H$ represents the case of \emph{unlabelled trajectories} which will be fed to \ac{MaxEnt-IRL}.
To generate a dataset with \emph{labelled trajectories} $\hat{H}$ we calculate the score using $r(s)$ as shown in Section~\ref{sec:objectives}. $\hat{H}$ are the input to \ac{DM-IRL}.


At each iteration, we sample the following datasets reflecting different types of history of user interactions: $\hat{H}$, $H_{Optimal}$, $H_{SubOptimal-0.2-MB}$,  $H_{SubOptimal-0.6-MB}$, $H_{SubOptimal-0.2-NB}$, $H_{SubOptimal-0.6-NB}$  each of size $15,000$ and $|\zeta_i|$ $\in [30, 40]$.

\noindent
\textbf{Evaluation process.}
To evaluate the performance of \ac{ISO}, we report the expected state value under optimal policy (\statevalue) for an \emph{initial} \is and an \emph{optimized} one, which we derive after around 100 iterations.
A higher expected state value means users are more satisfied while interacting with the \is. We initialize $40$ different initial \iss by randomly sampling reward functions and transition distribution, and report the overall performance over these $40$ systems. The true reward functions and the connectivity graphs of these sampled systems are fixed in the whole optimizing process.

\section{Results and Discussion}
\label{sec:results}

\noindent

\begin{figure*}[ht]
\centering
\vspace{-2mm}
   \includegraphics[clip, width=\textwidth]{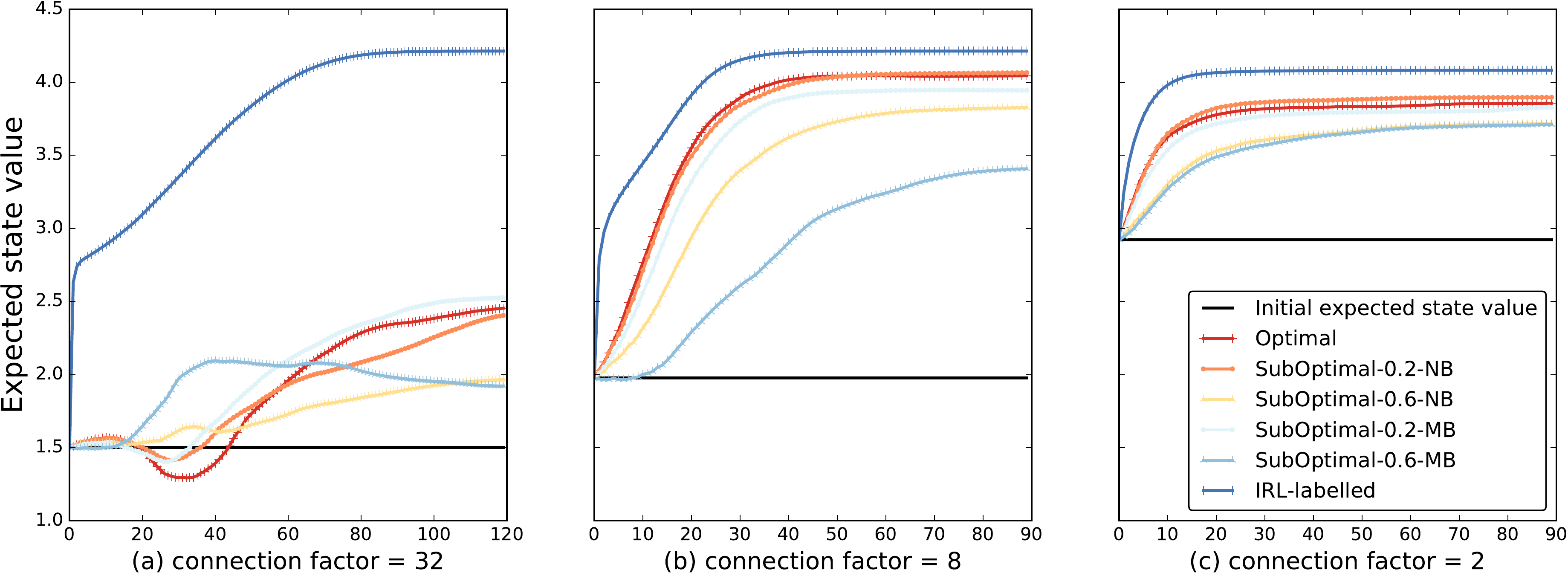}
   \caption{Performance of ISO over 40 randomly sampled systems. The x-axis is the number of iterations  of ISO and the y-axis is the expected state value.}
   \label{fig:state_value}
   \vspace*{-0.5\baselineskip}
\vspace{-2mm}   
\end{figure*}

\noindent
\textbf{Improving interactive systems with ISO.}
Figure~\ref{fig:state_value} shows how the quality of the interactive system increases with each iteration of \ac{ISO} in terms of different connection factors. 
As expected, when the user gives feedback about the quality of the trajectories (IRL-labelled), the task is simpler and \ac{ISO} manages to get high improvements with the oracle rewards. 
However, the picture changes when we hide the labels from the trajectories.  
Without labels, \ac{ISO} relies on the optimality of user behavior to recover the reward function. 
As the optimality decreases, so does the behavior of \ac{ISO}, and the performance decays.
With the oracle rewards from \ac{DM-IRL}, \ac{ISO} converges quite fast -- as we can see in Figure~\ref{fig:state_value}(b) and Figure~\ref{fig:state_value}(c), after 20 iterations the expected state value begins to plateau. Most improvements happen in the first several iterations. 
Thus, \ac{ISO} works with accurately labeled trajectories, but usually obtaining high-quality labels is intractable and expensive in a real \is because the real rewards are invisible. We report it as the oracle performance in our experiment.

With respect to unlabelled trajectories ISO is able to improve the initial expected state value. 
In Figure~\ref{fig:state_value}(a), the influence of the noise factor and types of trajectories (MB or NB) is clear. However, in Figure~\ref{fig:state_value}(c) where there are fewer connections between two states, only the convergence speeds of different curves are different but they all converge to the same state value eventually. \ac{ISO} manages to optimize the \is even though the user trajectories are quite noisy. 

More remarkable, the convergence speed and final converged values are different depending on the connection factors. As we can see, it is more difficult to get high performance when there are more connections between different states in the predefined systems. More connections mean that more possible trajectories could be taken and it is intractable for \ac{MaxEnt-IRL} to learn a reward function from this kind of situation. By contrast, in Figure~\ref{fig:state_value}(c), each state-action pair can only have two possible next states and the final average state value is much higher than the system in Figure~\ref{fig:state_value}(a). 


\noindent
\textbf{Impact of \ac{ISO} components.}
The performance of \ac{ISO} depends on its two components: (1)~\ac{RL} methods used to optimize the user policy $\pi$ for the original \ac{MDP} and system policy $\pi^+$ for transformed MDP$^{+}$; and (2)~\ac{IRL} methods -- to recover the true reward function.
The dependence on \ac{RL} methods is obvious -- the end result will only be as good as the quality of the final optimization, so an appropriate method should be used.
The performance of \ac{ISO} can be influenced by the quality of the recovered reward functions, $r(s)$.
For the case of labeled trajectories, the values of $r(s)$ recovered by \ac{DM-IRL} are identical to the ground truth since a regression model is used and we have the exact label for each user trajectory.
%
For the case of unlabelled trajectories, the quality of the recovered reward function is worse than \ac{DM-IRL}. \ac{MaxEnt-IRL} can only give a general overview of $r(s)$ if the user trajectories are optimal. If there are not enough constraints on the connections between states, with each iteration of running \ac{ISO}, the shape of the sampled trajectories becomes more similar, which means that most trajectories pass by the same states and the diversity of trajectories decreases. 
We found that this makes it even more difficult to recover $r(s)$ and the \ac{MaxEnt-IRL} quality deteriorates with the number of iterations, which results in lower performance in  Figure~\ref{fig:state_value}(a).
Since we are using one of the most common \ac{IRL} methods in this work, more advanced \ac{IRL} methods could be adopted to achieve higher performance. Hence, improving the performance of \ac{IRL} methods is likely to significantly boost the performance of \ac{ISO}.

\section{Conclusions and Future work}
\label{section:conclusion}
We have recognized that previous work on \iss has relied on numerous assumptions about user preferences.
As a result, \iss have been optimized for manually designed objectives that do not align with the true user preferences and cannot be generalized across different domains.
To overcome this discrepancy, we have proposed a novel algorithm: the \acf{ISO}, that both infers the user objective from their interactions, and optimizes the \is according to this inferred objective. 

Firstly, we model user interactions using \ac{MDP}, where the agent is the user, and the stochastic environment is the \is. User satisfaction is modeled by rewards received from interactions, and the user interaction history is represented by a set of trajectories.
Secondly, we infer the user needs from the observed interactions, in the form of a data-driven objective. Importantly, our method works without any domain knowledge, and is thus even applicable when prior knowledge is absent.
Thirdly, \ac{ISO} iterates between optimizing the \is for the current inferred objective; and letting the user adapt to the new system behavior.
This process repeats until both the user and system policies converge. Our experimental results show that \ac{ISO} robustly improves the user experience across different types of user behavior. 

Since optimizing an \is based on data-driven objectives is novel, many promising directions for future work are possible.
For instance, while \ac{ISO} performs well for users with a single goal, this approach could be extended to settings with multiple goals.
Similarly, extensions considering more personalized goals could benefit the overall user experience.
Finally, investigating the scalability and real world applicability of \ac{ISO} could open many research possibilities.

\setlength{\bibsep}{6pt plus 0.3ex}
\bibliographystyle{nips}
\bibliography{bibliography}

\clearpage


\end{document}